\documentclass[a4paper,twoside,12pt]{article}

\usepackage[utf8]{inputenc}
\usepackage[T1]{fontenc}
\usepackage[a4paper,twoside,top=2.5cm,bottom=4cm,inner=3cm,outer=2cm,footskip=2cm]{geometry}
\usepackage{setspace}
\usepackage{amsmath}
\usepackage{amssymb}
\usepackage{amsthm}
\usepackage[unicode,bookmarksopen]{hyperref}
\usepackage{pbox}
\usepackage{cite}

\begin{document}

\title{Generalizing k-means for an arbitrary distance matrix}
\author{Balázs Szalkai \\{\small Eötvös Loránd University, Budapest, Hungary}}
\maketitle

\begin{abstract}
The original k-means clustering method works only if the exact vectors representing the data points are known. Therefore calculating the distances from the centroids needs vector operations, since the average of abstract data points is undefined. Existing algorithms can be extended for those cases when the sole input is the distance matrix, and the exact representing vectors are unknown. This extension may be named {\em relational k-means} after a notation for a similar algorithm invented for fuzzy clustering. A method is then proposed for generalizing k-means for scenarios when the data points have absolutely no connection with a Euclidean space.
\end{abstract}

\section{Introduction}

The standard k-means method \cite{kmeans} takes a set of data points $p_1, ... p_n \in \mathbb{R}^d$ and a number of clusters $N$. Its aim is to produce an arrangement of the data points into $N$ clusters (that is, a labeling function $\ell : \{p_i\}_{i=1}^n \to \{1, ... N\}$) so that the following objective is minimized:

\[
\sum_{i = 1}^{n} ||p_i - z_i||^2,
\]

where $z_i = \frac{1}{|S_i|} \sum_{j \in S_i} p_j$, and $S_i = \{p_j : \ell(p_j) = \ell(p_i)\}$.

The main difficulty of this method is that it requires the data points to be the elements of a Euclidean space, since we need to average the data points somehow. In practice we often have data points (e.g., protein sequences) and a distance function which is not derived from some Euclidean representation. Even worse, the distance function may not be a metric at all. Clustering schemes like k-means are not applicable for these cases, as k-means requires vectors as input.

Various generalizations and extensions of k-means have been developed \cite{Cheung20032883} \cite{Dhillon:2004:KKS:1014052.1014118}, but none yet seems to have addressed the above problem. However, the fuzzy c-means clustering method is reported to have been successfully generalized \cite{Hathaway1994429}. The generalized method is known as Non-Euclidean Relational Fuzzy C-means {\em (NERF c-means)}. A similar extension of k-means, which can be viewed as a vast simplification of NERF c-means, is described in the next sections.

\section{Relational k-means}

Suppose first that we have a Euclidean distance matrix between the data points, but the exact location of the vectors representing them is unknown for us. Let $A \in \mathbb{R}^{n \times n}$ be the squared distance matrix, namely, $A_{ij} = ||p_i - p_j||^2$. Our objective is to calculate the squared norms $||p_i-z_i||^2$. The $p_i-z_i$ distance vectors are a special case of those linear combinations of the $p_j$ points where the sum of the coefficients is zero. That is, $p_i-z_i = \sum_{j=1}^n \lambda_j p_j$ for some suitable $\vec{\lambda} \in \mathbb{R}^n$, which satisfies the condition $\sum_{j=1}^n \lambda_j = 0$.

In fact, it can be easily verified that the squared length of $\sum_{i=1}^n \lambda_i p_i$ can be calculated by knowing only the matrix $A$:

\[
||\sum_{i=1}^n \lambda_i p_i||^2 = \sum_{i=1}^n \sum_{j=1}^n \lambda_i \lambda_j \left<p_i, p_j\right> = -\frac{1}{2} \sum_{i=1}^n \sum_{j=1}^n \lambda_i \lambda_j ||p_i - p_j||^2 = -\frac{1}{2}\lambda^\top A \lambda,
\]

In the above transformation we made use of the fact that $\sum_{i=1}^n \lambda_i = 0$.

Calculating a centroid distance is thus possible by computing a quadratic form. This means that, even if the only thing we know is the squared distance matrix $A$, we can run practically any k-means heuristic without substantial modifications. Of course, the time complexity will be impaired, as computing a quadratic form is an expensive operation.

\section{The non-Euclidean case}

Let $e_i$ denote the $i$th standard basis vector, and, for an index set $S \subset \{1, ... n\}$ let $\chi(S) := \sum_{i \in S} e_i$. Now let $z_S := \frac{1}{|S|}\sum_{j \in S} p_j$ denote a centroid. The formula $d^2(p_i, z_S) := -\frac{1}{2}\lambda^\top A \lambda$ (where $\lambda := \frac{1}{|S|}\chi(S) - e_i$) still makes sense even if $A$ has not been derived from Euclidean distances. Therefore, the above formula yields a generalization of the centroid distances.

This means that now we can speak of the weighted arithmetic mean of abstract data points in a sense that there is a possible interpretation of distance between two objects of that kind.

The above generalization shows that any k-means algorithm can be adapted to abstract distances. It is questionable though whether this generalized clustering method yields interesting and useful clusters. A completely arbitrary matrix $A$ can produce strange results. That is, if $-\frac{1}{2}\lambda^\top A \lambda$ takes a negative value for some vector $\lambda$ (the sum of whose coordinates is zero), then the distance defined by $\lambda$ will be negative. Of course, this is not possible in the case of a Euclidean squared distance matrix.

Negative distances can be eliminated by ensuring that $A_1$ is negative definite, where $A_1$ is the restriction of the quadratic form $A$ to the linear hyperplane perpendicular to $\vec{1}$. ($\vec{1}$ is the vector whose coordinates are all $1$.) This may require a modification to the original squared distance matrix, a modification that should be as small as possible in some sense.

A method proposed in \cite{Hathaway1994429} called $\beta$-spread transformation may be applicable here as well. That is, all the pairwise distances are gradually increased by the same amount until we have a matrix of the desired kind. This approach was reported to work well with fuzzy c-means for real-world data. The real-world suitability of an analogous matrix correction method for generalized k-means is yet to be evaluated.

\bibliography{generalized-kmeans}
\bibliographystyle{plain}

\end{document}